\RequirePackage{fix-cm}
\documentclass[twocolumn]{svjour3}          
\smartqed  
\usepackage[colorlinks,linkcolor=red, citecolor=blue]{hyperref}
\usepackage{graphicx}
\usepackage{cite}
\usepackage{subcaption}
\usepackage{color}
\usepackage{amssymb}
\usepackage[ruled]{algorithm2e} 
\usepackage{url}
\usepackage{multirow}
\usepackage{indentfirst} 
\usepackage{natbib}
\usepackage{amsmath,amssymb}
\begin{document}
\title{CDistNet: Perceiving Multi-Domain Character Distance

for Robust Text Recognition}

\author{
Tianlun Zheng \and Zhineng Chen* \and Shancheng Fang \and Hongtao Xie \and Yu-Gang Jiang
}

%
%
\institute{T. Zheng, Z. Chen (corresponding author) and Y.G. Jiang are with School of Computer Science and Shanghai Collaborative Innovation Center of Intelligent Visual Computing, Fudan Universtiy, Shanghai 200438, China. (E-mail: tlzheng21@m.fudan.edu.cn, zhinchen@fudan.edu.cn, ygj@fudan.edu.cn) \\
S. Fang and H. Xie are with School of Information Science and Technology, University of Science and Technology of China, Hefei 230026, China.(E-mail: fangsc@ustc.edu.cn, htxie@ustc.edu.cn)
}




\date{Received: date / Accepted: date}

\maketitle

\begin{abstract}

The Transformer-based encoder-decoder framework is becoming popular in scene text recognition, largely because it naturally integrates recognition clues from both visual and semantic domains. However, recent studies show that the two kinds of clues are not always well registered and therefore, feature and character might be misaligned in difficult text (e.g., with a rare shape). As a result, constraints such as character position are introduced to alleviate this problem. Despite certain success, visual and semantic are still separately modeled and they are merely loosely associated. In this paper, we propose a novel module called Multi-Domain Character Distance Perception (MDCDP) to establish a visually and semantically related position embedding. MDCDP uses the position embedding to query both visual and semantic features following the cross-attention mechanism. The two kinds of clues are fused into the position branch, generating a content-aware embedding that well perceives character spacing and orientation variants, character semantic affinities, and clues tying the two kinds of information. They are summarized as the multi-domain character distance. We develop CDistNet that stacks multiple MDCDPs to guide a gradually precise distance modeling. Thus, the feature-character alignment is well built even various recognition difficulties are presented. We verify CDistNet on ten challenging public datasets and two series of augmented datasets created by ourselves. The experiments demonstrate that CDistNet performs highly competitively. It not only ranks top-tier in standard benchmarks, but also outperforms recent popular methods by obvious margins on real and augmented datasets presenting severe text deformation, poor linguistic support, and rare character layouts. In addition, the visualization shows that CDistNet achieves proper information utilization in both visual and semantic domains. Our code is available at \url{https://github.com/simplify23/CDistNet}.


\end{abstract}


\maketitle
\section{Introduction}

\begin{figure*}[t]
\centering
\includegraphics[width=1\textwidth]{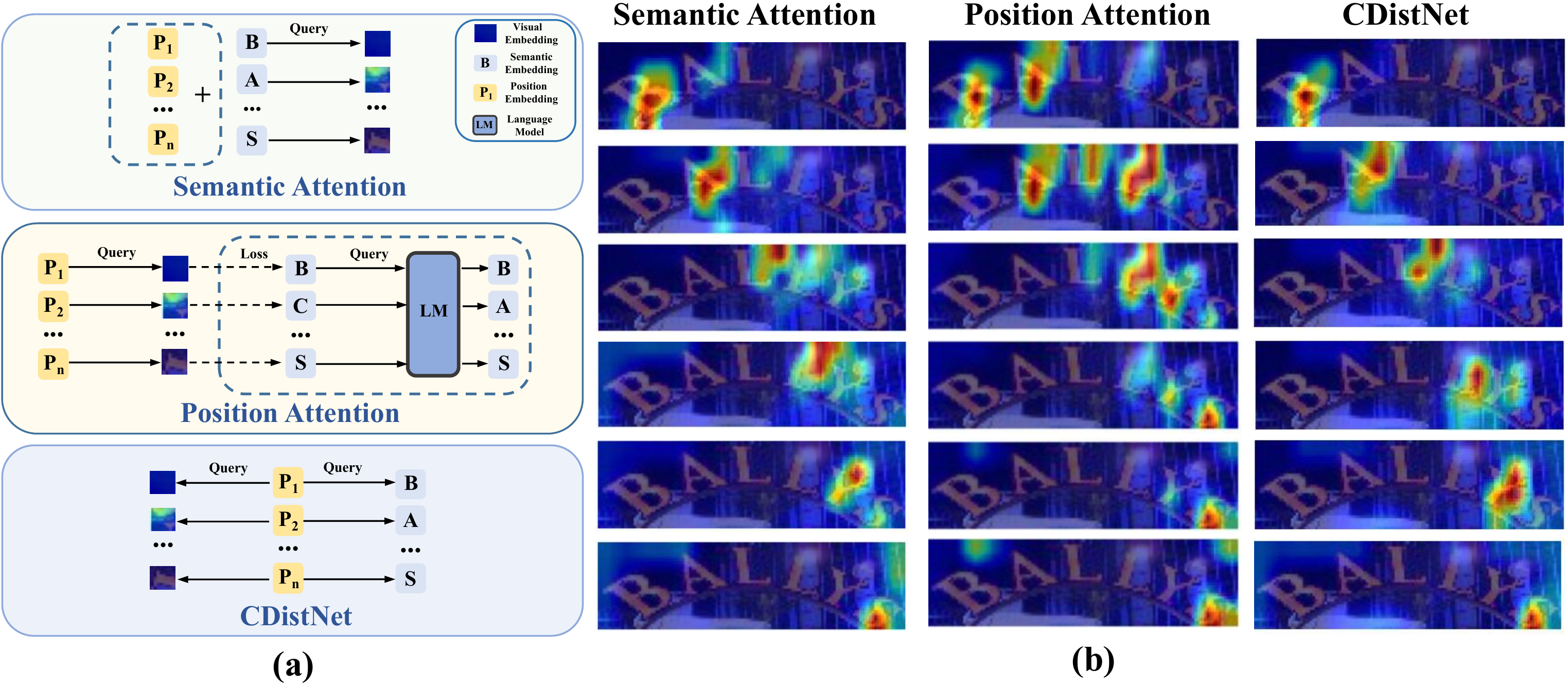} 
\caption{\textbf{Semantic Attention} \citep{wang2019simple_transformer,sheng2019nrtr,shi2018aster,li2019sar}: the semantic feature (optionally concatenated with the position embedding) is employed to query visual features for attention modeling. It easily causes \emph{attention drift}. (the first column in (b), line 3,4); \textbf{Position Attention} \citep{SRNyu2020towards,ABInet21CVPR,wang2021FTO,yue2020robustscanner}: a single position embedding is used to query visual features. The position attention solely may wrongly attend multiple characters in a single time step (the second column in (b), line 1,2,3,4). It is typically accompanied by semantic attention or successive position attention (on different domains) for joint visual and semantic modeling; \textbf{CDistNet}: the position embedding queries both semantic and visual features, forming a multi-domain tightly coupled feature and resulting in more proper attention localization (the third column in (b)). Solid and dotted lines denote the general forward pass and loss-guided computation, respectively. The dotted box means optional.}
\label{fig1:motivation}
\end{figure*}

Scene text recognition aims to read text in natural images. It has attracted wide interest in the computer vision community due to its pivotal role in extracting high-level textual information that is critical for many vision-related applications. Although extensive studies are carried out with significant progress over the past years \citep{ShiBY17crnn,ABInet21CVPR,Baekwhats_wrong_19ICCV,lee2016attention_origin,Du2022SVTR,zheng2023mrn}, the task still remains challenging for several difficulties, e.g., complex text deformations, unequally distributed characters, multiple fonts, cluttered backgrounds, etc. How to tackle these challenges is becoming a key issue for modern text recognizers. 

Recently, Transformer-based encoder-decoder methods have attained impressive performance in this task. For example, Sheng et al.~\citep{sheng2019nrtr} extract visual feature from the text image, and inject semantic feature from the text label to the decoder side. During decoding, the text image is identified character-by-character. At each time step, the semantic feature first concatenates with the sinusoidal position embedding. It is then used as the \emph{query} vector to align the corresponding character in the visual space, by which a character-level recognition decision is derived. The pipeline is summarized as the semantic attention scheme \citep{sheng2019nrtr,Baekwhats_wrong_19ICCV,SEEDqiao2020seed,zheng2023tps++}, as shown in the top part of Fig.\ref{fig1:motivation}(a). It has advantages such as providing a unified way of integrating recognition clues from visual and semantic domains. However, recent studies \citep{wang2020DAN,yue2020robustscanner} show that the two kinds of clues depend on each other. It is explained as when one of the two clues is weak, the other could not stably find its counterpart. Therefore, feature and character are easily mismatched in highly distorted text or rarely-seen character sequences. Moreover, it is also observed in \citep{yue2020robustscanner} that the mismatch is more prevalent in long text. Since the semantic feature is gradually reinforced and dominates the recognition with the accumulation of decoded characters, the position clue is weakened and features between neighbor time steps may become similar, easily causing the visual feature acquired by the model is not accurately aligned with the target character (see the first column of Fig.\ref{fig1:motivation}(b), line 3,4). The phenomenon is termed as \emph{attention drift} in the literature \citep{cheng2017focusing_attention,liao2019two_dim_per,wan2020textscanner,wang2019scene}.   
In view of the problem above, several studies \citep{wan2020textscanner,yue2020robustscanner,SRNyu2020towards,ABInet21CVPR} decouple position and semantic features by employing a single position branch as the query for more flexible attention catching (see the middle part of Fig.\ref{fig1:motivation}(a)). It generates the so-called position attention scheme. This scheme is typically accompanied by semantic attention or successive position attention on different domains for better feature utilization. For example, \citep{yue2020robustscanner} employs both a position attention on visual space and the semantic attention. The two kinds of attention are late fused to generate the decoded character. This method alleviates the mismatch somewhat and gets improved accuracy. Nevertheless, its position embedding is \emph{content-free}, where a character order-related fixed embedding is employed to associate with visual features in different text images. It is more like a placeholder constraint rather than a content-aware receptor, which cannot represent the diverse character patterns appearing in a fixed position and generates sub-optimal alignments. We observe that it may wrongly attend multiple characters in a single time step, as shown in Fig.\ref{fig1:motivation}(b) (the second column, line 1,2,3,4). To tackle this, besides modeling the position embedding to query visual feature at first. \citep{ABInet21CVPR} further uses the generated position embedding for a side-loss guided semantic modeling, followed by a query on the language model (LM), which injects the semantic knowledge, and further endows the model with the vocabulary-based rectification capability. It further improves the recognition accuracy to some extent. However, the side loss-guided semantic modeling is loosely associated with the visual space. The position utility is still underestimated by such a usage. Meanwhile, endowing the semantic information using external LMs may degrade the performance in cases no such model is readily available.

We argue that more robust feature-character alignments could be reached by using the position embedding to query both semantic and visual features in parallel, which is barely considered in existing studies. Such a scheme gives an embedding that not only perceives recognition clues from both sides. It also enjoys the merit that correlations between the two domains are simultaneously modeled. In other words, it also models the associations between 2D image space (i.e., character spacing and orientation variants) and 1D sequence space (i.e., character semantic affinities), utilizing them as a whole to get more robust features. Compared to modeling them sequentially \citep{SRNyu2020towards,ABInet21CVPR} or using the late fusion \citep{yue2020robustscanner}, it gives a more complete feature description thus more accurate feature-character alignment. As a result, it leads to robust recognition especially in challenging scenarios. Since this process can be vividly understood as describing the character correlation in and between visual and semantic domains. We call it multi-domain character distance in this paper.

Motivated by the observation above, in this paper we develop a novel module called \emph{Multi-Domain Character Distance Perception} (MDCDP) to characterize the multi-domain character distance. Similar to existing studies \citep{vaswani2017transformer, yue2020robustscanner}, MDCDP is initialized with a placeholder-like fixed position embedding. But differently, the embedding is used to \emph{query} both visual and semantic features following the cross-attention mechanism (see the bottom part of Fig.\ref{fig1:motivation}(a)). Therefore, it does not only impose position enhancement on the visual domain, but also delineates the character dependence in the semantic domain. By merging features from both domains, it generates a new multi-domain tightly coupled position embedding that well perceives spatial and semantic distance among characters. In addition, the embedding is then used as the \emph{query} vector of the next MDCDP, to obtain a more attention-focused feature-character alignment. Following this idea, we propose a novel architecture named CDistNet that stacks MDCDP several times to guide a gradually precise multi-domain character distance modeling. As seen in the third column of Fig.\ref{fig1:motivation}(b), CDistNet more accurately localizes the characters. It ensures that, at each time step, the correct feature is utilized for character decoding even complex character spatial variations are presented, therefore benefiting the recognition. Note that CDistNet achieves the recognition without using any external LM.

We conduct comprehensive experiments to evaluate CDistNet on ten public datasets, and two series of augmented datasets with increasing horizontal and curved deformation. On the public datasets, CDistNet consistently ranks top-tier on six standard regular and irregular text benchmarks. Meanwhile, it also outperforms recent popular methods on venues presenting severe text deformation and poor linguistic support, indicating that it is robust towards various text irregularities and less depends on external LMs. While on the augmented datasets, CDistNet attains greater advantages as the deformation intensity level rise when compared to recent popular methods, showing its superiority in handling text with rare character layouts observed at high-intensity levels. We also visualize the visual attention heatmap and the semantic affinity matrix during decoding. It is shown that not only a proper visual attention localization is obtained, but also a more complete character-involved semantic inference is observed. The results clearly demonstrate the effectiveness of CDistNet as well as the merit of modeling the multi-domain character distance.

Contribution of this paper is threefold. First, we propose MDCDP that employs the position feature to \emph{query} both visual and semantic features. In addition to exploring recognition clues from both domains, it obtains clues that tie the two kinds of information together, forming a comprehensive feature representation that deeply perceives the target character. Second, we develop CDistNet, a novel scene text recognizer. It stacks multiple MDCDPs to well perceive the multi-domain character distance and builds robust feature-character alignments, benefiting the recognition from a variety of difficulties. Third, we conduct extensive experiments and compare CDistNet with existing leading methods. It achieves highly competitive results and shows advantages in fighting against severe text deformation, poor linguistic support, and rare character layouts. In addition, the visualization experiments imply that CDistNet achieves proper information utilization in both visual and semantic domains.

\section{Related Work}
Scene text recognition is a long-standing computer vision task intensively studied over decades. Comprehensive surveys can be found in~\citep{ye2014text,long2021str_era,Chen2021text}. Recently, research efforts mainly lie in the irregular text recognition while sequence-based methods become popular~\citep{bhunia2021jvsr,masktextspotterlyu2018mask,jaderberg2016reading,rodriguez2015label,Baekwhats_wrong_19ICCV,cheng2018aon,nguyen2021dictionary,luo2021separating}, as different recognition clues (e.g., visual, semantic) can be jointly modeled to get the prediction. Based on how these clues are utilized, we can broadly categorize them into semantic-free, semantic-attention, and position-attention methods.

\subsection{Semantic-free Methods}
Sequence-based methods advance traditional character split-based methods~\citep{bai2014chinese,jaderberg2016reading} in avoiding the difficult individual character splitting problem, while allowing relationships between local image regions as well as semantic information to be utilized. Earlier methods in this branch are semantic-free methods. They implement the recognition by exploring image visual features mainly, while the semantic relationship among characters is not explicitly modeled. For example, Shi et al. \citep{ShiBY17crnn} proposed the CTC-based method, where the visual feature extracted by CNN was reshaped as a feature sequence and then modeled by RNN and CTC loss. Following this pipeline, several methods were developed with improved accuracy by developing a deep-text recurrent network~\citep{he2016reading}, ensembling multiple RNNs~\citep{su2017accurate}, devising a graph convolutional network-guided CTC decoding~\citep{hu2020gtc}, using the ViT-like encoder~\citep{Du2022SVTR}, etc. Rather than decoding by RNN, segmentation-based methods \citep{liao2019two_dim_per,xing2019convolutional,li2017fcsem}regarded the recognition as a segmentation problem where each character is a target category.  However, they generally required character-level annotations which were not always readily available. While they were also sensitive to segmentation noise. In sum, the performance of semantic-free methods was limited, partly due to not directly modeling recognition clues from other domains.

\subsection{Semantic-attention Methods}
Semantic-attention methods~\citep{chen2019adaptive,cluo2019moran,shi2018aster,sheng2019nrtr} utilize the semantic clue to reinforce the visual feature and generally use the attention-based encoder-decoder framework. Lee et al.~\citep{lee2016attention_origin} first introduced the attention mechanism into scene text recognition. It employed both the 1D image feature sequence and the embedding of character sequence, by which the semantic clue was leveraged. Later, it was extended by designing a more natural 2D image feature \citep{li2019sar}, adding dedicated attention enhancement modules \citep{cheng2017focusing_attention}, etc. By substituting RNN with Transformer \citep{vaswani2017transformer}, a powerful model in capturing global dependence of 1D sequence, variants \citep{sheng2019nrtr,wang2019simple_transformer,2dattentionlyu20192d} were developed with improved accuracy. Note that the position encoding was commonly applied to Transformer-based models whose attention mechanism was non-local. For example, sinusoidal position embedding was employed to record the character position in \citep{sheng2019nrtr}. The embedding is then appended to the semantic feature vector as auxiliary information. Later, learnable embedding was developed to adaptively utilize the position clue in \citep{devlin2019bert,lan2019albert}. Despite great processes made, it was observed that misalignment between feature and target character (i.e., local image region) still existed especially in long text, i.e., \emph{attention drift} \citep{cheng2017focusing_attention,liao2019two_dim_per,yue2020robustscanner,wang2019scene}.

\begin{figure*}[t]
\centering
\includegraphics[width=1.0\textwidth]{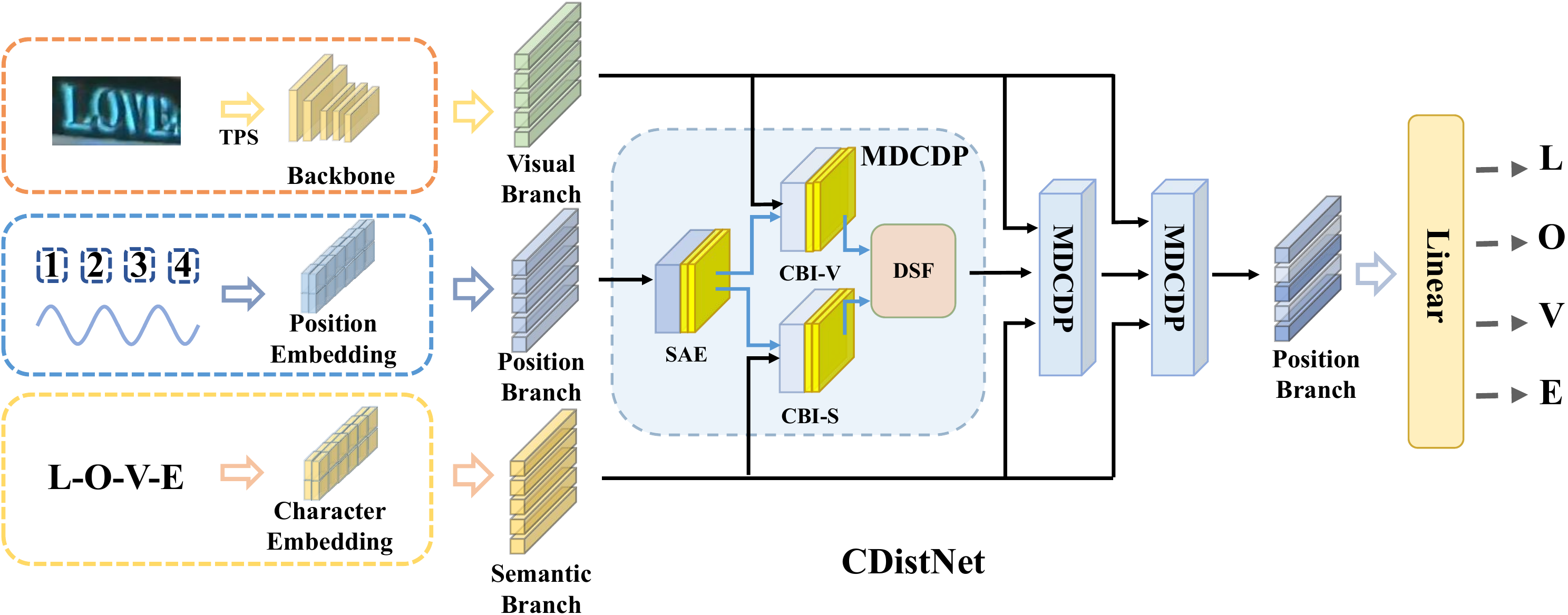} 
\caption{An overview of the proposed CDistNet. The encoder consists of visual, semantic and position branches. The three kinds of features are fed into the \emph{Multi-Domain Character Distance Perception} (MDCDP) module that learns a multi-domain tightly coupled feature representation. SAE, CBI and DSF are the proposed blocks. CDistNet stacks three MDCDPs to get a gradually precise distance modeling. Output of the last MDCDP is leveraged to decode the characters sequentially.}

\label{fig2}
\end{figure*}

\subsection{Position-attention Methods}
Position-attention methods emphasize developing a dedicated character position branch to ease the recognition, which has been taken into account by many recent methods. To suppress \emph{attention drift}, TextScanner \citep{wan2020textscanner} developed a position order segmentation map to ensure that characters were read in the right order and separated properly. RobustScanner \citep{yue2020robustscanner} devised a dedicated position enhancement branch along with a dynamic fusion mechanism. It achieved impressive results on contextless text. Meanwhile, several parallel decoding efforts were proposed by introducing semantic loss of different forms \citep{SEEDqiao2020seed, wang2021FTO, ABInet21CVPR}. For example, a placeholder sequence was initialized and learned to describe the character positions. Since LM can leverage vocabulary to rectify character-level recognition errors, some studies also focused on better utilizing this clue for performance improvement. They include: enabling flexible character-level correction \citep{da2022levenocr}, enriching the language representation \citep{wang2022petr,bautista2022parseq,wang2022mgp,he2021S-GTR}. \citep{xie2022wordart} also used position embedding to explore the character contour. Our work falls into the category of position-guided attention-based sequential decoding. However, different from \citep{yue2020robustscanner} that the position embedding only queries the visual feature, and \citep{ABInet21CVPR} that uses the position embedding to first query visual feature and then LM, we use the position embedding to query visual and semantic features and then fuse both queried features back to the position embedding, which also learns associations between the two domains. Our work thoroughly integrates the recognition clues in and between visual and semantic spaces. Therefore, it is less dependent on the LM and makes the recognition robust to difficult scenarios.

\section{Methodology}
The proposed CDistNet is illustrated in Fig.\ref{fig2}. It is an end-to-end trainable network that falls into the Transformer-based encoder-decoder framework. Specifically, the encoder consists of three branches, respectively for encoding visual, position and semantic information. On the decoder side, the three kinds of information are fused by a dedicated designed MDCDP module, in which the position branch is first leveraged to query both visual and semantic features. Then they are fused to generate a reinforced embedding, which describes the distance between characters in both visual and semantic domains. It is used as the position embedding of the next MDCDP to guide a more accurate distance modeling. The MDCDP is stacked several times in CDistNet to achieve a precise feature-character alignment. At last, characters are decoded sequentially based on the output of the last MDCDP.     

\subsection{Encoder}

\noindent\textbf{Visual Branch.}
The visual branch utilizes Thin-plate-splines (TPS) as a preprocess to rectify the text image \citep{shi2018aster,Baekwhats_wrong_19ICCV}. Then, similar to \citep{ABInet21CVPR,SRNyu2020towards}, ResNet-45 and Transformer units are employed as the backbone, which captures both local and global feature dependencies. The Transformer unit is a three-layer self-attention encoder with 1024 hidden units per layer. The process can be described using the following formula.

\begin{equation}
\begin{aligned}
    \textbf{F}_{vis} = \tau(\mathcal{R}( \mathbb{T}(\emph{I})))\in{R}^{N \times P\times E}
\end{aligned}
\end{equation}
where $I\in{R}^{W \times H}$ denotes the input text image, $\mathbb{T}$ is the TPS block. $\mathcal{R}$ denotes ResNet-45 and $\tau$ is the Transformer units. $N$ and $E$ are batchsize and the channels of visual feature, respectively. $P=\frac{WH}{64}$ is the length of the reshaped visual feature. Both the width and height are shrunk to 1/8 of the original size. $P$ and $E$ are also viewed as the length of visual feature sequence and the dimension of each feature, respectively. $E$ is empirically set to 512 for all three branches, i.e., they have the same feature dimension.

\noindent\textbf{Semantic Branch.}
Similar to \citep{sheng2019nrtr,yue2020robustscanner}, the semantic branch encodes the character labels. Since CDistNet is an autoregressive model that decodes the characters one-by-one, the semantic feature sequence is modeled differently for training and inference. In training, since all the character labels are already known, the feature sequence, i.e., $\textbf{F}_{sem}\in R^{N\times T\times E }$, is generated directly, where $T$ is a predefined parameter denoting the allowed maximum character number of a text instance, $E$ is the feature dimension in the sequence. Each feature is obtained by using Word2vec to transform the character into a vector. While in inference, since the character number is unknown in advance, the semantic embedding is updated step-by-step. Specifically, a fixed start token $\textbf{F}^0_{sem}\in R^{N\times 1\times E}$ is defined as the semantic embedding at time step 0. It is passed to the decoder side to predict the first character. With the just decoded character, we generate a new embedding $\textbf{F}^1_{sem}\in R^{N\times 2\times E}$ on the encoder side, where embedding of the character is appended. Then $\textbf{F}^1_{sem}$ is applied to decode the next character. Following this scheme, $\textbf{F}^t_{sem}\in R^{N\times (t+1)\times E}$ is obtained after decoding the $t$-th character. The whole process is terminated when the end token has been decoded.

\noindent\textbf{Position Branch.}
Similar to the semantic branch, the position branch encodes feature in one-time during training while the feature is updated step-by-step during inference. To encode the character positions in training, we first generate a feature sequence, each feature with a fixed constant $1/L$ in its position-indexed dimension while 0 otherwise, where $L$ is the length of the given text. Then, the same sinusoidal position embedding as in \citep{vaswani2017transformer} is applied, followed by two MLP layers to get the embedding $\textbf{F}_{pos}\in R^{N\times T\times E}$. While in inference, the position embedding is initialized with a placeholder $\textbf{F}^1_{pos}\in R^{N\times 1\times E}$ as described in training but with a constant $1$ as the nonzero element. When the first character is decoded, we re-initialize the position embedding $\textbf{F}^2_{pos}\in R^{N\times 2\times E}$ similarly, but with a constant $1/2$ on the corresponding dimensions of the two vectors. Following this scheme, the size of position embedding grows with the accumulation of decoded characters, where $\textbf{F}^t_{pos}\in R^{N\times t \times E}$ defines the embedding of the $t$-th time steps, and $1/t$ is the nonzero elements when initialization. Note that the position embedding has no start and end token.

Note that the three branches are generated individually at the encoder side. For training, they are generated separately in one-time with no information exchange. For inference, at each time step, the semantic feature is expanded by the just decoded character. However only a visual-irrelevant character embedding is appended. Whereas the position feature is just a step-related re-initialization.

\subsection{MDCDP}
\begin{figure*}[t]
\centering
\includegraphics[width=1\textwidth]{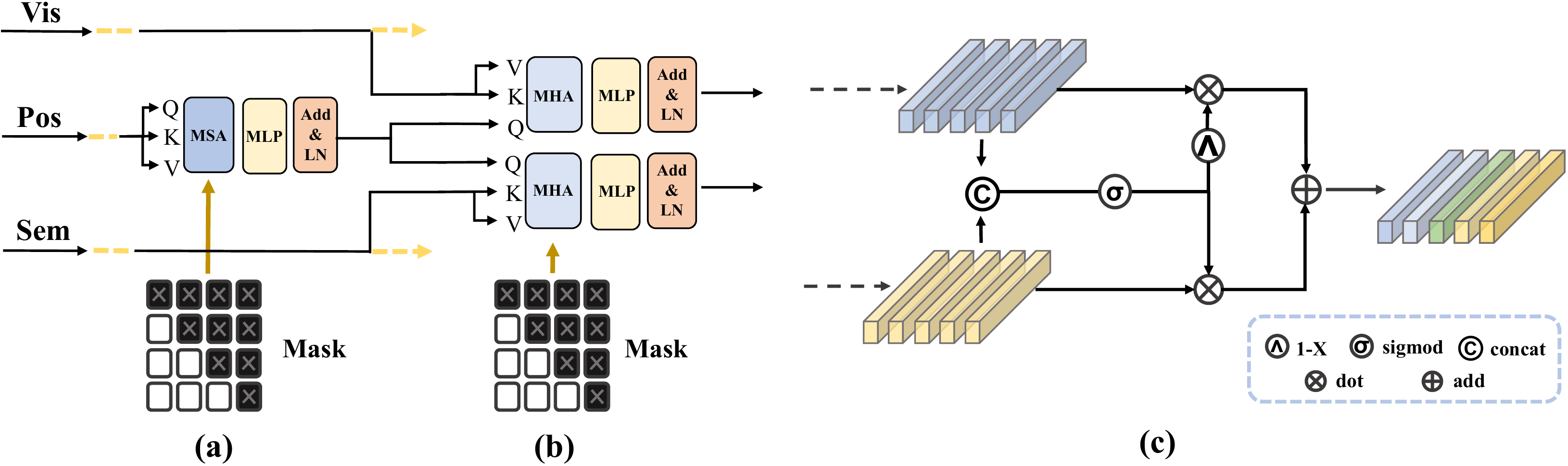} 
\caption{Details of the MDCDP module. (a) Self-Attention Enhancement (SAE), (b) Cross-Branch Interaction (CBI) and (c) Dynamic Shared Fusion (DSF).}
\label{fig:mdcdp block}
\end{figure*}

We now explain the formulation of MDCDP on the decoder side. MDCDP consists of three parts, a self-attention enhancement (SAE) for position feature reinforcement, a cross-branch interaction (CBI) that utilizes position feature to \emph{query} both the visual and semantic branches, obtaining position-enhanced representations, and a dynamic shared fusion (DSF) to get a visual and semantic joint embedding. As a result, the two features are fused and a multi-domain tightly coupled representation is obtained. The detailed structure of SAE, CBI and DSF are shown in Fig.\ref{fig:mdcdp block}, from which we can get an intuitive understanding of how \emph{Query} (Q), \emph{Key} (K) and \emph{Value} (V) are applied.

\noindent\textbf{Self-Attention Enhancement (SAE).}
At time step $t$, with the position embedding $\textbf{F}^t_{pos}$ from the encoder, SAE is applied at first to reinforce the embedding using one multi-head attention block \citep{vaswani2017transformer} but half its dimension to reduce the computational cost. In addition, an upper triangular mask $\textbf{M}^t_{pos}$ is applied to the query vector to prevent it from "seeing itself", or saying, leaking information across time steps. It happens in the second and subsequent MDCDPs, where the position embedding has already been mixed with visual and semantic information. Formally, this block is described as:
\begin{align}
    \textbf{F}^t_{pos} = Atten([\textbf{F}^t_{pos},\textbf{F}^t_{pos}],\textbf{M}^t_{pos})+FFN
\label{equ:sae}
\end{align}
\noindent where \emph{FFN} represents the feed forward network. \emph{Atten} is the multi-head attention. The first term in the square bracket is Q, while the second term represents both K and V. In this case it is multi-head self-attention.

Note that previous related studies \citep{ABInet21CVPR,SRNyu2020towards} did not consider such a self-attention-based position reinforcement. We argue that SAE can enable a more targeted position embedding through gradient back-propagation. It leads to a more flexible feature learning, especially for the second and subsequent MDCDPs. 

\noindent\textbf{Cross-Branch Interaction (CBI).} 
The enhanced position embedding $\textbf{F}^t_{pos}$ is then treated as the query vector and fed into the visual and semantic branches in parallel. When applied to the semantic branch, we call it CBI-S, and $\textbf{F}^t_{pos\_sem}$, the cross-attention between position and semantic feature is calculated. It simulates the semantic affinity between previously decoded characters and the character to be predicted. Similarly, $\textbf{M}^t_{pos}$ is leveraged to prevent the semantic information leaking across time steps. Meanwhile, $\textbf{F}^t_{pos\_vis}$, the cross-attention between position and visual feature, is also generated on the CBI-V side. It is explained as using the previously decoded character positions to search for the character region to be recognized in the text image. Thus, both branches are strengthened after the interactions. Note that visual feature of the whole image is readily available such that no mask is required. The two cross-attention blocks are defined as:
\begin{gather}
    \textbf{F}^t_{pos\_sem} = Atten([\textbf{F}^t_{pos}, \textbf{F}^t_{sem}],\textbf{M}^t_{pos})+FFN
\label{equ:cbi-s}
\\
    \textbf{F}^t_{pos\_vis} = Atten([\textbf{F}^t_{pos}, \textbf{F}_{vis}])+FFN
\label{equ:cbi-v}
\end{gather}

Note that previous studies \citep{Baekwhats_wrong_19ICCV,li2019sar} used semantic feature as the query to interact with visual feature, i.e., the semantic attention scheme. RobustScanner \citep{yue2020robustscanner} extended it by adding an additional query from position feature to visual feature and enables position-enhanced decoding, i.e., the position-attention scheme. However, the position query does not interact with the semantic branch. In contrast, we formulate the interactions as the position-based enhancement to both visual and semantic domains. It reinforces not only visual but also semantic features and can be understood as delineating both spatial variants and semantic affinities among characters.

\noindent\textbf{Dynamic Shared Fusion (DSF).}
DSF aims to fuse the two position-enhanced features and generate a multi-domain tightly coupled feature representation. It takes the two features as input. They are concatenated to form a hybrid feature whose channels are doubled. Then, the feature undergoes a $1\times1$ convolution that encourages feature fusion while halves the channels, i.e., retaining the same channels as either input feature. After that, a gating mechanism is designed to transform it to weight matrices $\hat{\textbf{S}}$, which are applied to both visual and semantic features element-wisely, forming a dynamic fusion of features across visual and semantic domains. Formally,
\begin{gather}
\mathrm{\hat{\textbf{S}} = \sigma([\textbf{F}^t_{pos\_sem}, \textbf{F}^t_{pos\_vis}]\textbf{W}_{conv})}
\\
\mathrm{\textbf{F}^t_{out} =   \hat{\textbf{S}}\otimes \textbf{F}^t_{pos\_sem}+(1-\hat{\textbf{S}})\otimes \textbf{F}^t_{pos\_vis}}
\label{eq_sf}
\end{gather}
where $\sigma$ is the sigmoid function, $\textbf{W}_{conv}\in\mathbb{R}^{2C\times C}$ denotes the corresponding convolution. $\otimes$ is the element-wisely dot product. Note that the fusion is efficient and similar blocks are also considered in \citep{ABInet21CVPR,SRNyu2020towards}.

Note that each MDCDP starts with the SAE block and ends with the DSF block. The two blocks perform a self-attention-based feature reinforcement and a gated-attention-based feature reinforcement, respectively. The two kinds of attention improve the feature representation from different but complementary aspects. We believe this complementary attention usage is also a key towards effective feature learning. They are in favor of tying recognition clues from both visual and semantic domains together, and hence, leading to robust recognition especially in challenging scenarios. In addition, one peculiarity of DSF is that $\textbf{W}_{conv}$, the convolutional parameters, are shared across time steps and among MDCDP modules. As a consequence, DSF not only decouples the feature fusion with the previous cross-attention-based modeling, but also eases the feature learning. We will verify its effectiveness in the ablation study.

\subsection{CDistNet}
We then explain the construction of an effective text recognizer based on the proposed MDCDP. Through DSF, we get a multi-domain tightly coupled representation that contains recognition-related visual and semantic clues. To enable thorough feature interaction, a stacked structure with several MDCDPs appended one-by-one is developed, where the number of stacked MDCDPs is empirically set to 3 (will be discussed in experiments). The structure constructs a gradually precise feature-character alignment by allowing the relevant features interactions thoroughly in and between visual and semantic feature spaces. Formally, it regards $\textbf{F}^{t,1}_{out}$, the output of the first MDCDP at time step $t$, as the position embedding of the next MDCDP, then sequentially generating $\textbf{F}^{t,2}_{out}$ and $\textbf{F}^{t,3}_{out}$. At last, a linear classifier with softmax activation is applied to $\textbf{F}^{t,3}_{out}$ to get the prediction, i.e., the $t$-th decoded character.

\begin{figure*}[t]
\centering
\includegraphics[width=1\textwidth]{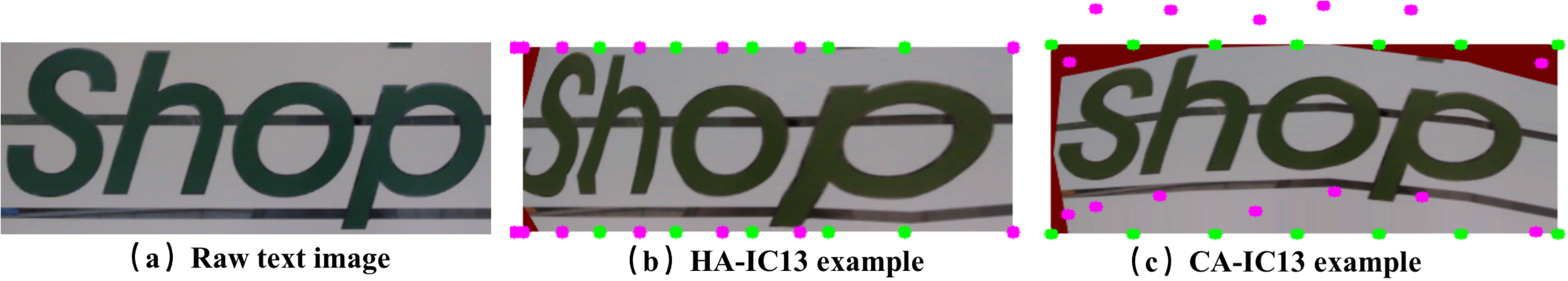} 
\caption{Raw text image and its augmentations. Green and purple points represent the initial fiducial points and their moved counterparts. (b) and (c) are the horizontal and curved stretching instances, respectively. Their deformation intensity levels are both set to 6, the most severe one. }
\label{fig:HA&CA_process}
\end{figure*}

\section{Dataset}

We first introduce the datasets employed to evaluate CDistNet. They include public datasets and augmented datasets constructed by ourselves. For the latter a subjective assessment is also conducted.

\subsection{Public Datasets}
CDistNet is evaluated on ten public real-world datasets. Their training and testing protocols are summarized as the following three types. First, for the most widely evaluated six standard benchmarks, all models are trained on two large-scale synthetic datasets and tested on these benchmarks. The datasets are as follows.

\noindent\textbf{MJSynth (MJ)} \citep{MJor90K} and \textbf{SynthText (ST)} \citep{ST} are two synthetic datasets each with millions of text images. 8.91M text instances from MJ and 6.95M from ST are retained for model training. 

\noindent\textbf{ICDAR2013 (IC13)}
is cropped from 288 real scene images. Following previous work~\citep{Baekwhats_wrong_19ICCV}, The version with 857 images is chosen for testing, which deletes non-alphanumeric characters and text instances shorter than 3 characters. 

\noindent\textbf{Street View Text (SVT)} \citep{SVT} 
contains 257 images for training and 647 images for testing, which are captured by Google Street View. The testing images are chosen for our experiments.

\noindent\textbf{IIIT5k-Words (IIIT5k)} \citep{IIIT5K} 
consists of 3000 testing images, which are collected from Google Image Searches. These instances are almost horizontal.

\noindent\textbf{ICDAR2015 (IC15)} \citep{ICDAR2015} 
is an irregular dataset with 1811 images for testing. Words in this dataset are mostly curved, perspective and/or shading, which are hard to recognize.

\noindent\textbf{SVT-Perspective (SVTP)} \citep{SVT-P} 
is also created from Google Street View. It contains 645 curved text images. 

\noindent\textbf{CUTE80 (CT80)} \citep{CUTE80} 
contains 288 irregular text images cropped from natural scene images.

Second, for larger and more difficult text datasets, i.e., ICDAR ArT \citep{chng2019art} and Uber-Text \citep{zhang2017uber} in our experiments, we first merged their training sets for model training from scratch. Then the models are tested individually on their test sets. Details of the two datasets are:

\noindent\textbf{ICDAR ArT (ArT)} \citep{chng2019art}
includes 32k training and 35k test text instances. It is created for recognizing arbitrary-shaped text. Thus, all types of text shapes, including horizontal, multi-oriented, and curved, are presented in this dataset. 

\noindent\textbf{Uber-Text (Uber)} \citep{zhang2017uber} 
contains street-level images collected from car-mounted sensors. It contains over 110k images with 4.84 text instances per image on average. The instances cover business names, street names and street numbers. We use the version processed by \cite{baek2021realdataset}, where vertical text has been excluded and only English text is reserved. It contains 128k training and 80k test instances, respectively. Most of them exhibit challenges like perspective, blurring, curved and multi-oriented, etc.

Third, for ICDAR MLT17 \citep{nayef2017mlt17} and ICDAR MLT19 \citep{nayef2019mlt19} with multilingual text, we train the models based on their own training set from scratch. Then the models are tested individually on their test sets 
The datasets are:

\noindent\textbf{ICDAR MLT17 (MLT17)} \citep{nayef2017mlt17} is designed for multilingual text recognition and contains text instances from seven languages: Arabic, Latin, Chinese, Japanese, Korean, Bangla, and symbols. There are nearly 6.7k training and 1.6k test instances in total. Text instances are presented in multiple scripts, font styles, and orientations, providing a challenging scenario for text recognition systems.

\noindent\textbf{ICDAR MLT19 (MLT19)} \citep{nayef2019mlt19} is also designed to recognize multilingual text. It includes text instances from eight languages: Arabic, Latin, Chinese, Japanese, Korean, Bangla, Hindi, and symbols. Due to the unavailability of MLT19 test data, we randomly split its training data into new training and test sets according to 9:1. There are nearly 7.5k training and 0.8k test instances in total. It exhibits similar challenges as MLT17. Meanwhile, the two datasets share text instances in part.

\begin{figure*}[t]
\centering
\includegraphics[width=1\textwidth]{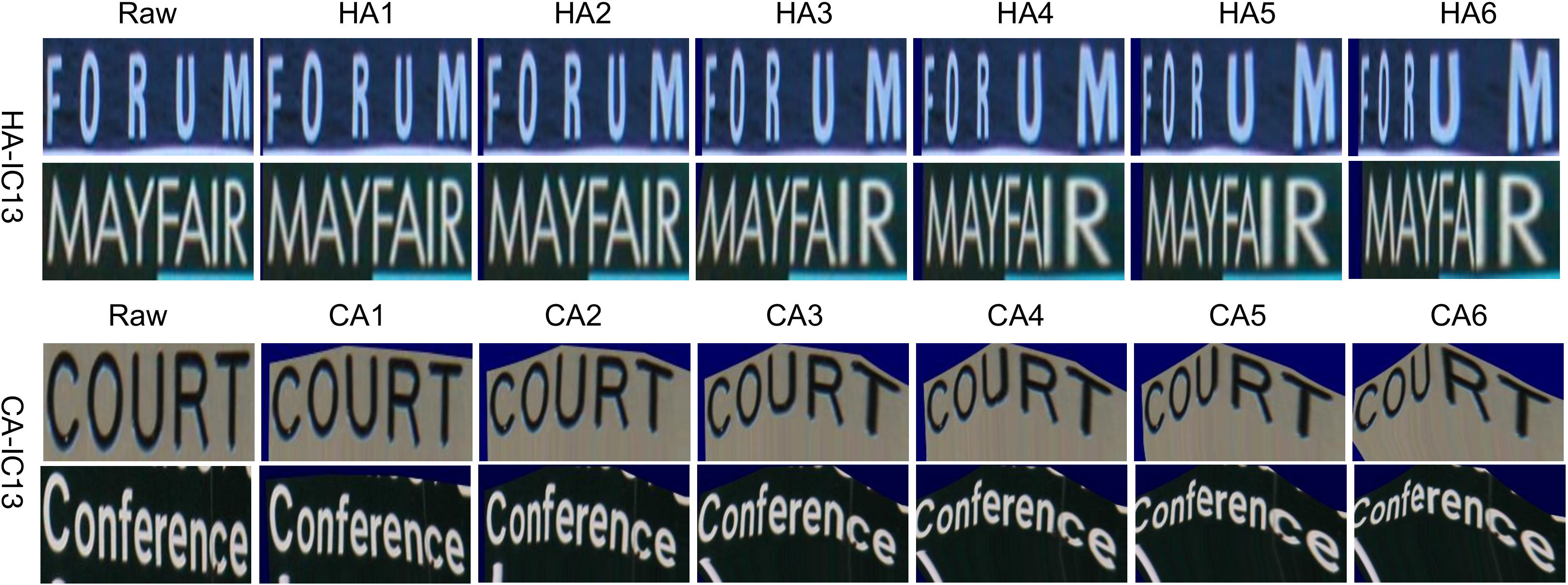} 
\caption{Augmented text images with different intensity levels in HA-IC13 (top) and CA-IC13 (bottom).}
\label{fig5:HA&CA_datasets}
\end{figure*}

\subsection{Augmented Datasets}
To thoroughly evaluate CDistNet especially for scenarios where character spatial layouts vary significantly, we construct two series of augmented datasets with different levels of intensity variants from the ICDAR2013 (IC13) dataset. Specifically, we employ horizontal and curved stretching of intensity from 1 (the smallest deformation) to 6 (the largest deformation) with an interval step of 1 to IC13 images. As a result, we obtain 12 counterparts of IC13 with different deformation levels. We group the datasets according to the stretching type and termed them as HA-IC13 (horizontal) and CA-IC3 (curved) series, respectively. For the name of each set, for example, HA5 denotes the set from HA-IC13 with intensity level 5. The rest is defined similarity. Each set simulates recognition difficulties of different levels. 

\begin{figure*}[]
\centering
\includegraphics[width=1\textwidth]{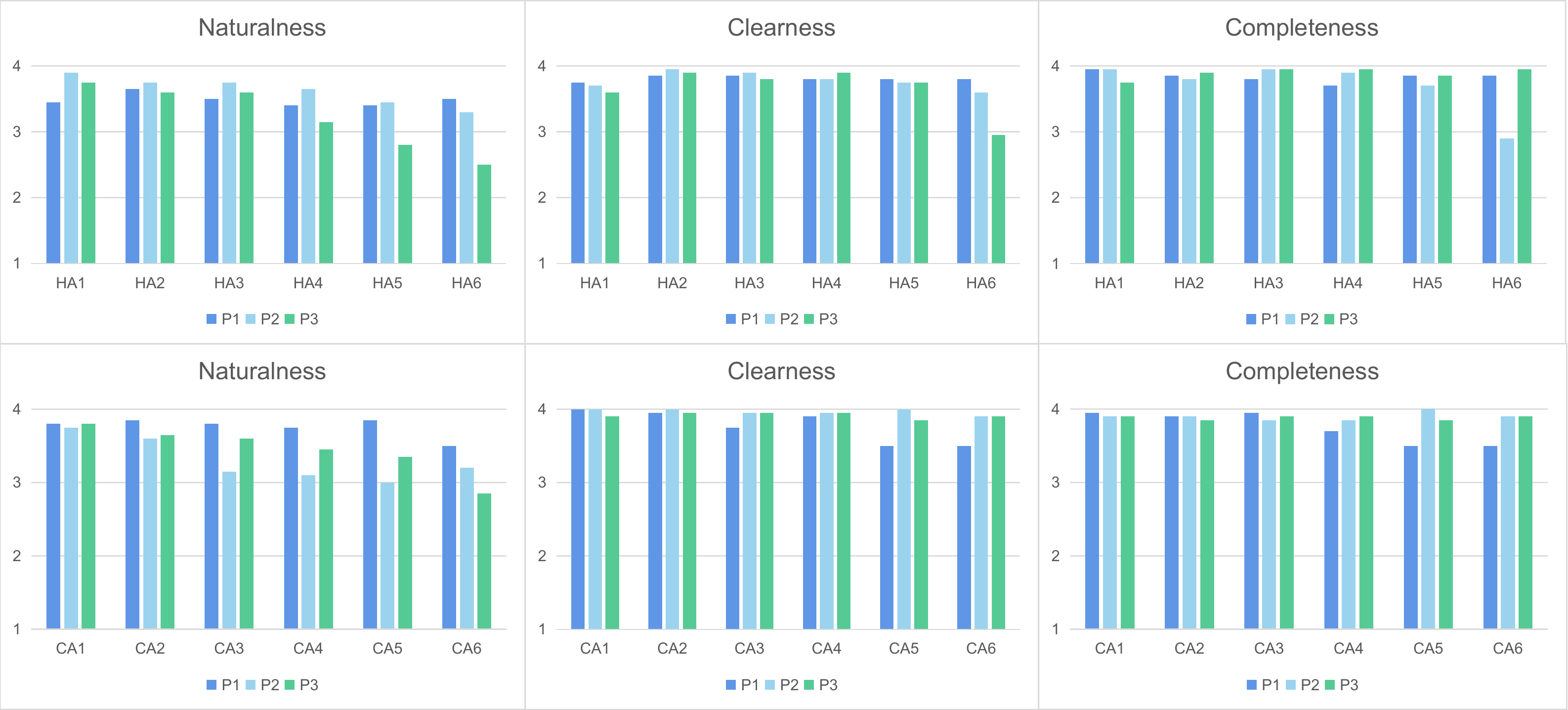} 
\caption{Subjective evaluation results of HA-IC13 and CA-IC13 datasets. Three participants are asked to assess the images from Naturalness, Clearness and Completeness independently.}
\label{human_eval}
\end{figure*}

We elaborate on the stretching details as follows. Given a text image $I\in{R}^{W \times H}$ as depicted in Fig.\ref{fig:HA&CA_process}(a), we first assign $2(N+1)$ fiducial points equally distributed along upper and lower boundaries of the image, i.e., the green points in Fig.\ref{fig:HA&CA_process}(b) and (c). An ${\frac{W}{4N} \times H}$ background region is also concatenated from the left side, as the first character may exceed the left boundary after the augmentation. Then, for each fiducial point $p_{i}=\left[l_{i}, h_{i}\right]$ with $h_{i}\in\{0, H\}$, we generate its horizontal and curved counterparts, i.e., the moved fiducial points defined as $p_{i}^{HA}$ and $p_{i}^{CA}$. They are calculated by using Equ.\ref{5} and Equ.\ref{6}, respectively.
\begin{gather}
p_{i}{ }^{H A}=\left[l_{i}+\theta_{M}, h_{i}\right] \label{5}\\
p_{i}^{C A}=\left[l_{i}+\theta_{M}, h_{i}-\theta_{M}\right] \label{6}
\end{gather}
where $\theta_{M}=\mu-\lambda*s$. $s\in \{1, 2, 3, 4, 5, 6\}$ is the deformation intensity level. $\mu \in \left[0, \frac{W}{4N}\right]$ is a random value. $\lambda$ is set to $max(\frac{W}{8N}, \mu)$, requiring $\lambda \geq \mu$ such that $\theta_{M}\leq 0$. That is to say, all fiducial points are moved to the left side in the X-axis (the horizontal case), and simultaneously the upper side in the Y-axis (the curved case), as the purple points shown in Fig.\ref{fig:HA&CA_process}(b) and (c). In the following, all image pixels are transformed according to the moved fiducial points using TPS transformation. With these operations, we create elastic deformations to the text image while roughly maintaining the geometric shape of each character.

In Fig.\ref{fig5:HA&CA_datasets} we list two examples undergoing different horizontal and curved stretching intensity levels. As can be seen, recognition difficulties such as text deformation, unequally distributed characters of different levels are successfully created. We term them to undergo varying character layouts. Note that the augmentation is an extension of the work of Luo et al.~\citep{luo2020learn_to_aug}. Compared to \citep{luo2020learn_to_aug}, the difference lies in that they focused on online augmentation and ignored constructing new datasets. While we narrow the allowed variants and create two series of augmented datasets paying attention to horizontal and curved deformations with different intensity levels. The datasets have been made publicly available along with the code.

\begin{table*}[h]
\centering
\caption{Ablation studies on SAE, CBI and DSF. Sem, Vis and Pos denote the semantic, visual and position branches, respectively. S\_V denotes using semantic feature to \emph{query} visual feature. The others are similarly defined. ($\checkmark$) means switching the roles of the two features. WS denotes weight sharing among MDCDP modules.}
\begin{tabular}{c|c|ccc|ccc|c|c c}
\hline
Ablation & \multirow{2}{*}{Line} &\multicolumn{3}{c|}{SAE} & \multicolumn{3}{c|}{CBI} & \multirow{2}{*}{Fusion}  & \multirow{2}{*}{SVT} & \multirow{2}{*}{IC15} \\
Part& & Sem & Vis & Pos & S\_V & P\_V & P\_S & & \\
\hline
\multirow{5}{*}{SAE} &1& & & & & $\checkmark$ & $\checkmark$ & DSF &92.27 & 84.92 \\
&2&$\checkmark$ &  & &  & $\checkmark$ & $\checkmark$ & DSF &92.89 & 84.70 \\
&3& & $\checkmark$ &  & & $\checkmark$ & $\checkmark$ & DSF & 93.04 & 84.65 \\
&4& $\checkmark$ & $\checkmark$ &  &  & $\checkmark$ & $\checkmark$ & DSF & 93.04 & 84.76 \\
&5& $\checkmark$ & $\checkmark$ & $\checkmark$ &  & $\checkmark$ & $\checkmark$ & DSF & 93.66 & 84.70 \\
\hline
\multirow{5}{*}{CBI}&6 &&& $\checkmark$ & & $\checkmark$ & & DSF & 90.26 & 82.11 \\
&7&& & $\checkmark$ & $\checkmark$ & & & DSF &91.34 & 84.04 \\
&8&& & $\checkmark$ & $\checkmark$ & & $\checkmark$ & DSF & 92.27 & 85.37 \\
&9&& & $\checkmark$ & $\checkmark$ & $\checkmark$ & & DSF & 92.89 & 84.93 \\
&10&& &$\checkmark$ & & $\checkmark$ & ($\checkmark$) & DSF & 93.35 & 84.70 \\
\hline
\multirow{3}{*}{DSF}&11&& & $\checkmark$ & & $\checkmark$ & $\checkmark$ & Add & 91.96 & 85.48 \\
&12&& & $\checkmark$ & & $\checkmark$ & $\checkmark$ & Dot & 91.81 & 84.21 \\
&13&& & $\checkmark$ & & $\checkmark$ & $\checkmark$ & DSF w/o WS & 93.51& 85.04 \\
\hline
&14&& & $\checkmark$ & & $\checkmark$ & $\checkmark$ & DSF & \textbf{93.66} & \textbf{85.92} \\
\hline
\end{tabular}
\label{table: Ablation study}
\end{table*}

\subsection{Subjective assessment on HA-IC13 and CA-IC13}
To verify whether the augmented HA-IC13 and CA-IC13 are suitable for evaluating scene text recognition models, we conduct user-based studies to rate the augmented text images. Specifically, we are concerned with whether the images are natural and legible. Therefore three rating dimensions are considered, i.e., naturalness, clearness, and completeness. Given a text image, naturalness quantifies its closeness level to real-world examples, clearness assesses its levels to be clearly read, completeness focuses on whether all characters are fully depicted. Note that the latter two dimensions together describe whether the image is legible.

For the assessment data, we randomly select 20 augmented images from each stretching type and intensity level. Therefore, 240 images are collected in total. Three participants, all graduate students working in computer vision topics, are recruited. They are asked to view the images one-by-one, and provide their assessments on the three dimensions independently. For each dimension, four options (i.e., Fully (4), Mostly (3), Partly (2), None (1)) representing the subjective agreement level are given. The participants are asked to choose one of them. To avoid bias, images with different intensity levels are mixed first and their levels are hidden when given to the participants. 

The evaluation results are illustrated in Fig.\ref{human_eval}, where the average score of each participant is plotted according to the evaluated dimensions and intensity levels. The main observations are as follows. 
\begin{itemize}
    \item For naturalness, all participants admit that it is perceptible that the images are rendered, and artificial levels of the images rise with the increase of intensity level. The most strict participant (i.e., P3) gives an average score between "Mostly" and "Partly" at HA5, HA6 and CA6. While the rest ratings are all above "Mostly", implying that the images appear basically natural except for a few severely deformed examples.   
    \item For clearness and completeness, all participants vote that the text is clearly and completely shown for most examples, indicating our augmentation basically keeps the readability of the text. The participants said although the text is severely deformed, the geometric shape of each character is largely maintained. So the text still remains recognizable to humans. For a few cases below "Mostly" (e.g., P2 in HA6), the participant said they are because of a few examples with characters partly out of bounds, and blurring caused by severe stretching.
\end{itemize}

After the generic evaluation above, we also talk with the participants about whether the images are suitable for evaluating scene text recognition models, and are the character layouts changed apparently and diversely. All participants give affirmative answers to the two questions. Therefore, we basically verify that both HA-IC13 and CA-IC13 are rendered natural and legible in most cases. They present diverse character layouts and are suitable for evaluating text recognizers.

\section{Experiments}
We introduce the implementation details at first, then present the ablation study, followed by experimental results and discussions grouped by datasets according to the major difficulties they presented.

\subsection{Implementation details}
We resize the text image to 32$\times$128 and employ the data augmentation in \citep{ABInet21CVPR}, i.e., image quality deterioration, color jitter and geometry transformation. The allowed maximum character number of a text instance, i.e., $T$, is set as 25 for the six standard benchmarks and two series of augmented datasets, and set as 60 for ArT, Uber, MLT17 and MLT19, as their instances typically involve more characters. To train CDistNet, the initial learning rate is set to 4$\times 10^{-4}$. The first 10k iterations use Warm-up. The whole training iterations are determined by  
\begin{equation}\label{lr}
lr = d_{model}^{-0.5}\cdot\min({n}^{-0.5},{n} \cdot {warm\_n}^{-1.5})
\end{equation}
where $n$ and $warm\_n$ denote the number of normal and Warm-up iterations. $d_{model}$ is set to 512. For the ablation study, all models are trained for 6 epochs. When compared with existing methods, CDistNet is trained for 8 epochs following Equ.\ref{lr} and then another 2 epochs with a constant learning rate of $10^{-5}$. The batch size is set to 700. For experiments on ArT, Uber, MLT17, MLT19, HA-IC13 and CA-IC13, the compared methods are all retrained by 10 epochs as described above for fair comparison. All models are trained using their codes directly except for S-GTR. S-GTR has not provided the code of its graph-based textual reasoning module. So we implement a similar graph neural network instead.

To reduce the computational cost of CDistNet, we shrink the transformer units employed, where the dimension of the MLP layer is reduced from 2048 to 1024 in the encoder, and from 2048 to 512 in the decoder. The number of encoder and decoder layers are both set to 3. Beam search is applied to determine the decoded character sequence and its size is empirically set to 10. All models are trained by using a server with 6 NVIDIA 3080 GPUs on PyTorch. 

\begin{table*}[h]
\centering
\caption{Accuracy comparison of different methods. * denotes our implementation with improved accuracy. Bracket values are accuracy improvements compared with results reported by their papers.}
\begin{tabular}{c|c|ccc}
\hline
Method & \#MDCDP & SVT &  IC15 & Speed (ms)\\
\hline
CDistNet & 1 & 92.74 &  84.82 & 61.48 \\
CDistNet & 2 & \textbf{94.28}	& 84.87 & 87.68\\
CDistNet & 3 & 93.66 &\textbf{85.92} & 123.28\\
CDistNet & 4 & 93.82 &	84.43 & 149.99\\
\hline
CDistNet w/o Sem & 3 &90.26 & 82.11 & 80.91\\
NRTR*\citep{sheng2019nrtr} & -- & 91.34 (-0.16) & 84.04 \textbf{(+5.00)} & 122.74 \\
RobustScanner*\citep{yue2020robustscanner} & -- & 91.96 \textbf{(+3.86)} & 84.21 \textbf{(+7.11)} & 122.78\\
\hline
\end{tabular}
\label{table:layer comparsion}
\end{table*}

\subsection{Ablation study}
To better understand the components of CDistNet, we carry out controlled experiments on both SVT (regular text) and IC15 (irregular text) under different configurations as follows.

\noindent\textbf{The effectiveness of SAE.} 

SAE imposes self-attention-based feature refinement on the applied branch. Besides the position branch, it is curious that the recognition would be improved by applying SAE to the visual and semantic features on the decoder side. With this doubt in mind, we apply SAE to different branches while Tab.\ref{table: Ablation study} gives the result. It is seen that the differences are marginal when equipping SAE to either or both visual and semantic branches (line 2,3,4). Nevertheless, on the position branch only the improvement is noticeable (line 14). It is better than applying SAE to none or all three branches (line 1,5). The result is in line with our expectations. The visual feature is extracted from a powerful hybrid backbone while the semantic feature is dynamically fortified during decoding. It is less meaningful to reinforce them further. In contrast, the position feature comes from a fixed embedding, while SAE allows dynamically reinforcing its feature, especially for the subsequent MDCDP modules. Therefore improvements are observed.

\noindent\textbf{The effectiveness of CBI.}
There are multiple ways to establish the CBI. We enumerate six of them and give the results in Tab.\ref{table: Ablation study} (line 6-10,14). CDistNet (line 14) achieves the best result among the competitors, better than the semantic attention scheme (e.g., NRTR-like implementation \citep{sheng2019nrtr}, line 7), the late fusion of semantic attention and position attention (e.g., RobustScanner-like implementation \citep{yue2020robustscanner}, line 9) and the scheme of using semantic feature to query position feature (S\_P, line 10). The results basically validate two observations. First, applying the position attention to visual feature is helpful (see line 7,9). Second, imposing the position attention on semantic feature is also effective (see line 6,10,14), which enables perceiving semantic affinities between previous decoded characters and the character to be recognized.

\noindent\textbf{The effectiveness of DSF}
We have tested four operations to fuse the position-enhanced visual and semantic features. As shown in Tab.\ref{table: Ablation study}, DSF outperforms static-based fusions (Add and Dot) (line 11,12) as well as the scheme that not sharing weights among MDCDP modules (line 13). The results show the effectiveness of the gated-attention-based fusion scheme, which is complementary to the SAE described above.

\noindent\textbf{The number of MDCDP modules considered.}
Stacking more MDCDPs tends to generate more comprehensive feature representation, but increases the computational cost. Ablation study on this point is given by Tab.\ref{table:layer comparsion} (the top half), where recognition accuracy and inference speed are both presented. The best accuracy is reported when three MDCDPs are equipped. It performs better than cases where fewer (one and two) or more (four) modules are considered. Therefore the number of MDCDP modules is set to 3 in CDistNet.

\begin{table*}[th]
\centering
\caption{Accuracy comparison with existing methods on six standard benchmarks.}
\begin{tabular}{|c|c|c|c|c|c|c|c|c|}
\hline
Method&Year&Traning Data&IC13&SVT&IIITK&IC15&SVTP&CT80\\
\hline
CRNN\citep{ShiBY17crnn} & 2017 & 90K  & 86.7 & 80.8 & 78.2 & -- & --& --  \\
FocusAtten\citep{cheng2017focusing_attention}& 2017 & 90K+ST & 93.3 & 85.9 & 87.4 & 70.6 & -- & --\\
AON\citep{cheng2018aon}& 2018 & 90K+ST  & -- & 82.8 & 87.0 & 68.2 &73.0 & 76.8\\
ASTER\citep{shi2018aster}& 2018 & 90K+ST  & 91.8& 89.5 & 93.4 & 76.1 & 78.5 & 79.5\\
ESIR\citep{zhan2019esir}& 2019 & 90K+ST  & 91.3 & 90.2 & 93.3 & 76.9 & 79.6 & 83.3 \\
SAR\citep{li2019sar}& 2019 & 90K+ST & 91.0 & 84.5 & 91.5 & 69.2 & 76.4 & 83.3\\
2D-Attention\citep{2dattentionlyu20192d} & 2019 & 90K+ST& 92.7 & 90.1 & 94.0 & 76.3 & 82.3 & 86.8 \\
MaskTextSpotter\citep{liao2019mask} & 2019 & 90K+ST & 95.3 & 90.6 & 93.9 & 77.3 & 82.2 & 87.8 \\
NRTR\citep{sheng2019nrtr} & 2019 & 90K+ST & 95.8 & 91.5 & 90.1 & 79.4 & 86.6 & 80.9\\
SE-ASTER \citep{SEEDqiao2020seed} & 2020 & 90K+ST  & 92.8 & 89.6 & 93.8 & 80.0 & 81.4 & 83.6 \\
Textscanner\citep{wan2020textscanner} & 2020 & 90K+ST & 92.9 & 90.1 & 93.9 & 79.4 & 84.3 & 83.3\\
DAN\citep{wang2020DAN} & 2020 & 90K+ST  & 93.9 & 89.2 & 94.3 & 74.5 & 80.0 & 84.4 \\
RobustScanner\citep{yue2020robustscanner} & 2020 & 90K+ST  & 94.8 & 88.1 & 95.3 & 77.1 & 79.5 & 90.3\\
SRN\citep{SRNyu2020towards} & 2020 & 90K+ST & 95.5 & 91.5 & 94.8 & 82.7 & 85.1 & 87.8\\
Pren2D\citep{yan2021primitive} & 2021 & 90K+ST+Real & 96.4& 94& 95.6& 83 & 87.6& 91.7\\
JVSR\citep{bhunia2021jvsr} & 2021 & 90K+ST & 95.5 & 92.2 & 95.2 & 84.0 & 85.7 & 89.7 \\
VisionLAN\citep{wang2021FTO}& 2021 & 90K+ST  & 95.7& 91.7& 95.8 & 83.7 & 86.0 & 88.5\\
ABINet\citep{ABInet21CVPR} & 2021 & 90K+ST & \textbf{97.4} &93.5& 96.2&\textbf{86.0}&\textbf{89.3}&89.2\\

S-GTR\citep{he2021S-GTR} & 2022 & 90K+ST  & 96.8	& \textbf{94.1}	& 95.8	& 84.6	& 87.9 &	92.3
\\
\hline
CDistNet w/o TPS & -- & 90K+ST  & \textbf{97.4} & 93.6 & \textbf{96.7} & 85.2 & 88.8 & 92.0\\
CDistNet & -- & 90K+ST  & \textbf{97.4} & 93.5 & 96.4 & \textbf{86.0} & 88.7 & \textbf{93.4}\\
\hline
\end{tabular}
\label{table:sota2}
\end{table*}

\noindent\textbf{Contribution of the position utilization.}
There are several works encoding the position information. CDistNet differs from them not only in position utilization but also in other aspects. To make a fair comparison, we modified these methods where other differences are mostly eliminated. The results are listed in Tab.\ref{table:layer comparsion} (the bottom half).

We explain how the methods are modified. \emph{CDistNet w/o Sem} denotes the semantic branch is removed from CDistNet thus the position feature only queries the visual feature (Tab.\ref{table: Ablation study}, line 6). The absence of modeling the semantic clue, despite faster, leads to a noticeable accuracy drop of over 3\% on average. It implies the importance of modeling the semantic feature. In \emph{NRTR*} (Tab.\ref{table: Ablation study}, line 7), we strengthen the raw NRTR \citep{sheng2019nrtr} using our visual encoder, from which 5\% accuracy gain in IC15 is observed due to a superior visual feature. Besides the visual encoding branch, \emph{RobustScanner*} uses our "CNN+Transformer" to replace its "CNN+LSTM" implementation at the encoder side. It improves the accuracy by 3.86\% and 7.11\% on SVT and IC15, respectively. Our modification leads to noticeable improvements to their raw implementations.

With the modifications, accuracy gains from our parallel position attention scheme can be quantitatively evaluated. CDistNet still attains nearly 1.7\% accuracy gains on average in both datasets compared with RobustScanner*, largely attributed to that P\_S (applying the position attention to semantic feature) is a superior cross-attention scheme compared to S\_V (i.e., the semantic attention).

\begin{figure*}[ht]
\centering
\includegraphics[width=0.8\textwidth]{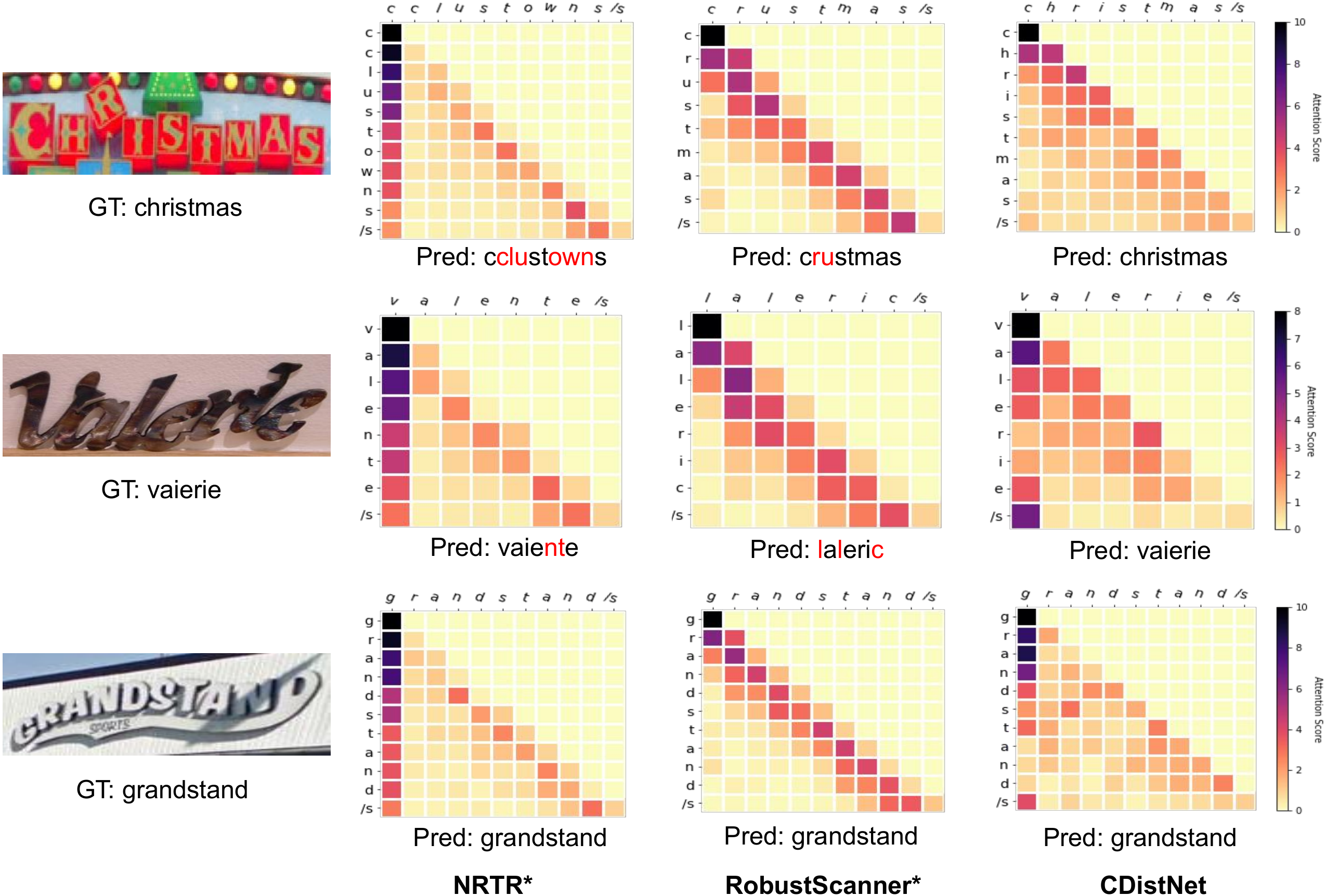} 
\caption{Visualization of the semantic affinity matrices of different methods. In each matrix a row represents the semantic affinities between previously decoded characters and the diagonal character. Darker color indicates a higher affinity value. The leftmost column is the text image and groundtruth (GT). Pred means the predicted result. Below defined similarity. The second to fourth columns are visualization of NRTR*, RobustScanner* and CDistNet. Red color means incorrectly recognized characters.}
\label{fig10:sem_pos_attention}
\end{figure*}

\subsection{Experiments on six standard benchmarks}
We compare CDistNet with nineteen existing methods published from year 2017 to 2022, covering the majority of influential methods of these years. The results are given by Tab.\ref{table:sota2}. It is shown that CDistNet achieves the best results on four datasets including IC13, IIITK, IC15 and CT80. As expected, the performance gains are more prominent in the irregular text datasets compared to regular text datasets, demonstrating its superiority in modeling multi-domain character distance thus benefiting the recognition of irregular text. We are also curious about whether the superior performance in the irregular text comes from the use of TPS block. Therefore, the model removing TPS block from CDistNet (i.e., \emph{CDistNet w/o TPS}) is also implemented. Its results on the three irregular text datasets, despite slightly lower, are still highly competitive. Interestingly, CDistNet w/o TPS even performs better than CDistNet in the three regular text datasets, implying that TPS block also produces a few undesired rectifications. The observation reveals that TPS block helps the recognition to some extent in challenging cases. However, the remarkable recognition capability is mostly attributed to the proposed CDistNet itself.

When comparing with previous state-of-the-art methods such as ABINet and S-GTR, CDistNet does not show obvious accuracy differences (i.e., within 1\% in five of the six datasets) although consistently ranks top-tier. It indicates that the performance of scene text recognition is approaching saturated on these benchmarks. Nevertheless, when looking into the closely related sequential decoding methods \citep{sheng2019nrtr,wan2020textscanner,yue2020robustscanner}, CDistNet outperforms them by large margins on all six datasets. It again demonstrates the superiority of the proposed feature modeling. It incorporates visual, semantic and position clues and well model fine-grained character distance, thus, better aligning visual feature with the target character even with various recognition difficulties.

\begin{figure*}[ht]
\centering
\includegraphics[width=0.8\textwidth]{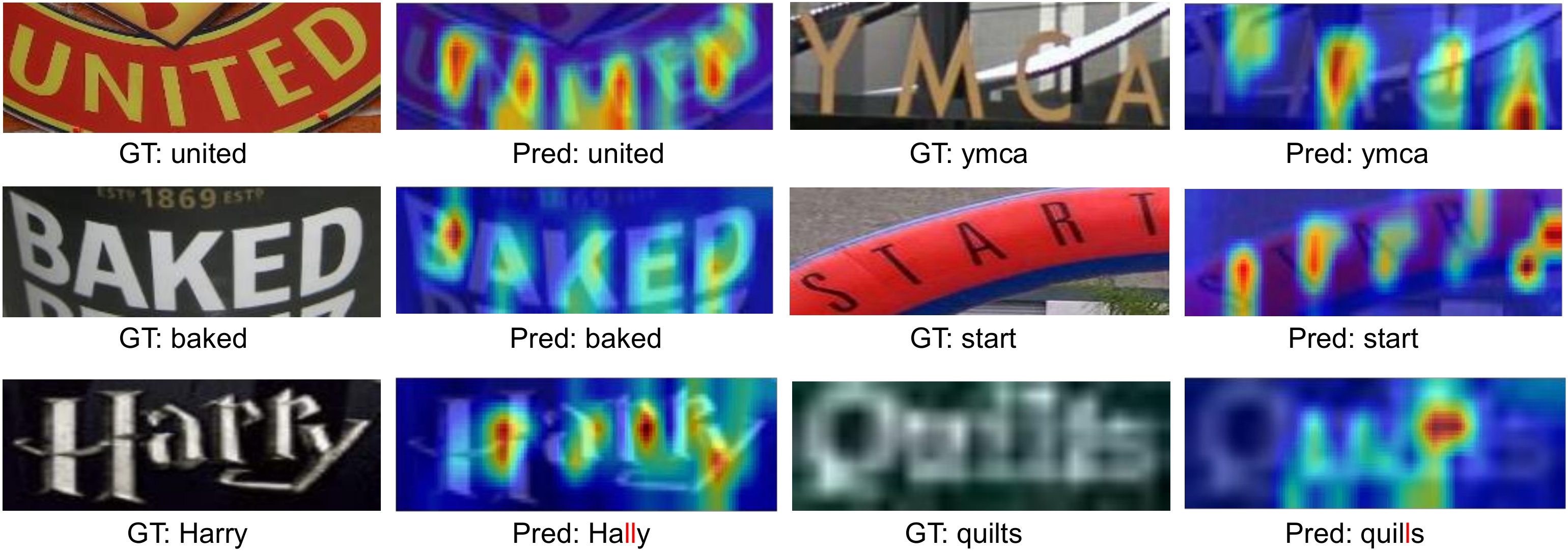} 
\caption{Attention map visualization of the position-enhanced visual feature in the last MDCDP module. Red color means incorrectly recognized characters.}
\label{fig:vis_case}
\end{figure*}

To better explore the semantic feature usage, we visualize three illustrative examples in the form of semantic affinity matrices in Fig.\ref{fig10:sem_pos_attention}, where the semantic utilization of three different methods (i.e., NRTR*, RobustScanner* and CDistNet) during the decoding are shown. For CDistNet, it is obtained by inspecting the position-enhanced semantic branch of the last MDCDP. While for the other two, they are the last self-attention block from the semantic branch.

Specifically, in each matrix, Y-axis denotes the decoding time steps while X-axis represents the decoded characters. Darker color indicates a higher affinity value, i.e., a higher contribution to recognizing the diagonal character. Two observations are obtained from the matrices. First, the diagonal elements play more important roles (darker) in CDistNet than in the other two methods. Since CDistNet jointly models recognition clues in and between visual and semantic spaces, the feature is more aware of the character to be recognized. It is thus more confident in semantic utilization. Second, for CDistNet the majority of previous decoded characters have darker colors thus more deeply contributing to the decoding process. It produces a more accurate prediction in general. On the contrary, in NRTR* the position and semantic features are not well decoupled. So the decoding mostly relies on the first column. While in RobustScanner*, although the position is decoupled, the decoding considers characters far from the recognized one less. It implies in part why the two methods sometimes make incorrect recognition. Conversely, it again demonstrates that CDistNet achieves a more comprehensive semantic affinity utilization.

\begin{table*}[]
\centering
\caption{Accuracy comparison of different methods on ArT, Uber, MLT17 and MLT19.}
\begin{tabular}{c|cc|cc}
\hline
Method   & ArT            & Uber           & MLT17         & MLT19         \\ \hline
S-GTR \citep{he2021S-GTR}      & 52.05          & 50.12          & 63.73         & 63.17         \\
ABINet \citep{ABInet21CVPR}   & 53.68          & 52.17          & 63.81         & 62.22         \\ \hline
CDistNet & \textbf{58.82} & \textbf{54.86} & \textbf{67.10} & \textbf{66.70} \\ \hline
\end{tabular}
\label{tab:art_uber_mlt}
\end{table*}

\begin{figure*}[]
\centering
\includegraphics[width=1\textwidth]{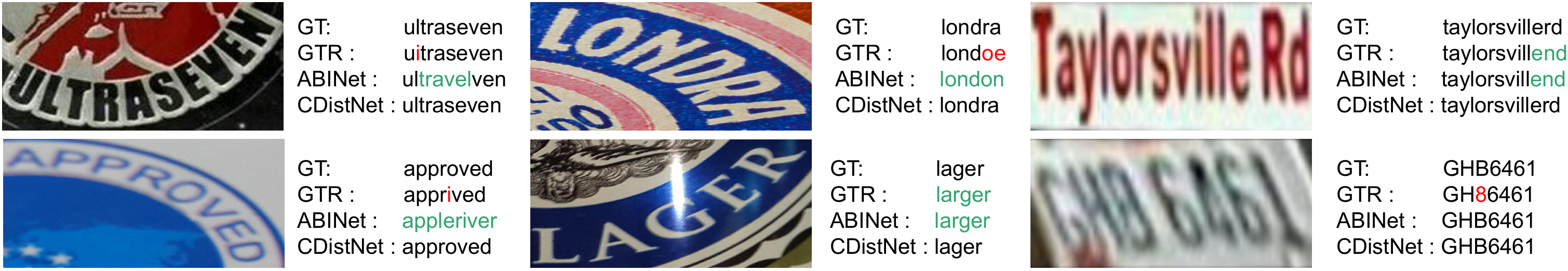} 
\caption{Recognition examples from ArT (the left two columns) and Uber-Text (the right column) datasets. Incorrect recognition relevant to erroneous vocabulary rectification is marked in green, others are marked in red.}
\label{vis_art}
\end{figure*}

\begin{figure*}[]
\centering
\includegraphics[width=1\textwidth]{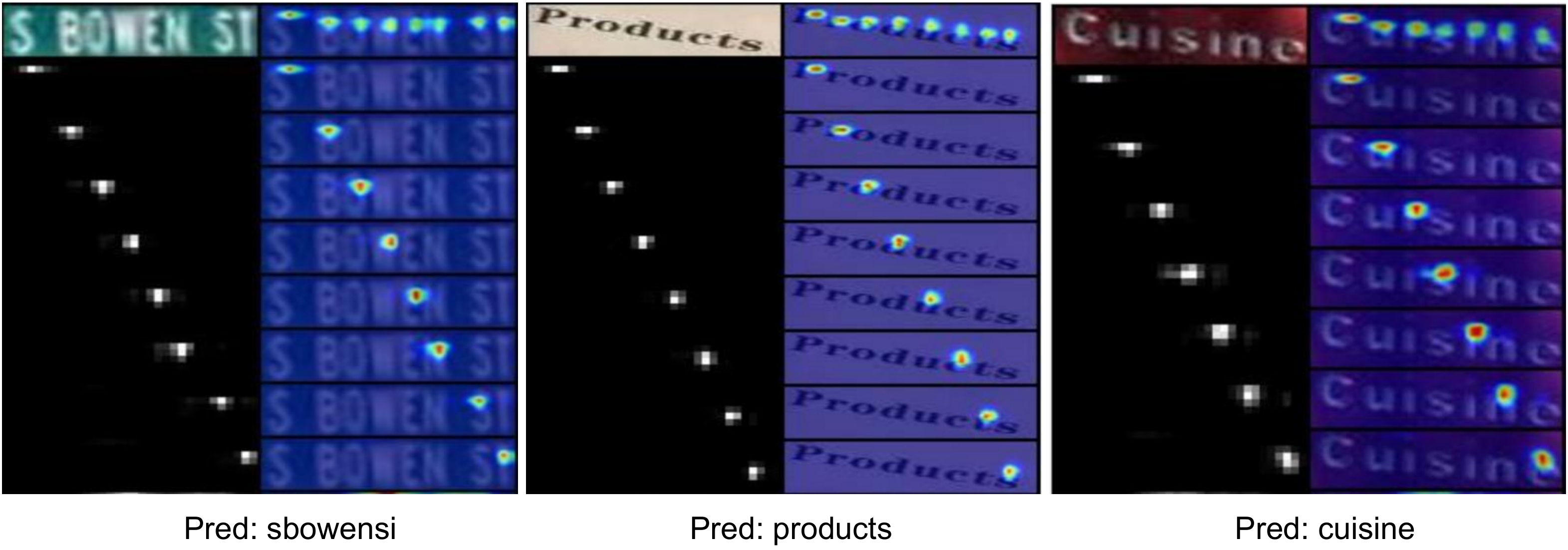} 
\caption{Step-by-step CDistNet visualization on ArT (the left two instances) and Uber-Text (the right one). The heatmaps are visualized based on the position embedding after the third MDCDP.}
\label{vis_char}
\end{figure*}

We then visualize the attention map based on the position-enhanced visual branch of the last MDCDP, where six examples are shown in Fig.\ref{fig:vis_case}. The characters are properly localized on the attention maps in most cases. It implies that CDistNet retains powerful and universal recognition capability with the existence of various difficulties. The multi-domain feature representation well encodes the spatial layout and semantic affinity among characters, thus ensuring that accurate feature-character alignments are established in most cases. On the other hand, it is observed that there are also a few wrongly recognized cases. The failures can be summarized as three categories mainly, i.e., multiple text fonts (e.g., harry), severe blur (e.g., quilts) and vertical text. Most of them are even indistinguishable to humans and they are still common challenges for modern text recognizers.

\subsection{Experiments on ArT and Uber}
As stated, performance on the six standard benchmarks is near saturated and cannot well distinguish the characteristic of different methods. Therefore we further evaluate CDistNet and compare it with ABINet and S-GTR, two methods with very close performance previously. The experiments are conducted on ArT and Uber at first. The two datasets contain an obviously larger number of text instances and present severe text deformation that includes high-intensity bending, perspective, blurring, etc.

The results on ArT and Uber are presented in the left half of Table \ref{tab:art_uber_mlt}. First, the accuracy for all three methods is not high, revealing the significant recognition difficulty of the two datasets compared to previous ones. Second, CDistNet gains obvious accuracy advantages compared to S-GTR and ABINet. It outperforms S-GTR by \%6.77 and ABINet by \%5.14 (corresponding to 13\% and 9.6\% relative improvements) on ArT, and S-GTR by \%4.74 and ABINet by \%2.69 (corresponding to 9.5\% and 5.2\% relative improvements) on Uber. 

\begin{figure*}[]
\centering
\includegraphics[width=1\textwidth]{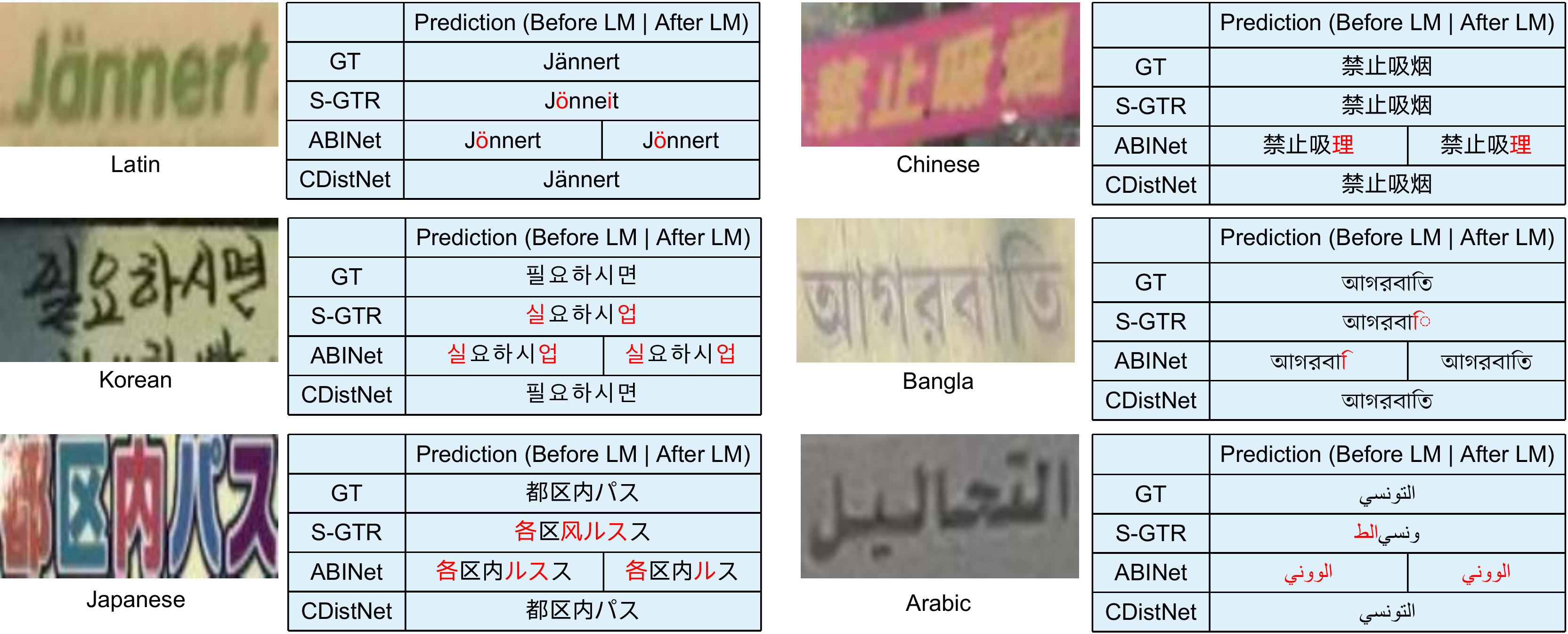} 
\caption{Recognition examples from MLT17 and MLT19 datasets. LM means language model. Incorrectly recognized characters are marked in red.}
\label{vis_MLT}
\end{figure*}

\begin{figure*}[]
\centering
\includegraphics[width=1\textwidth]{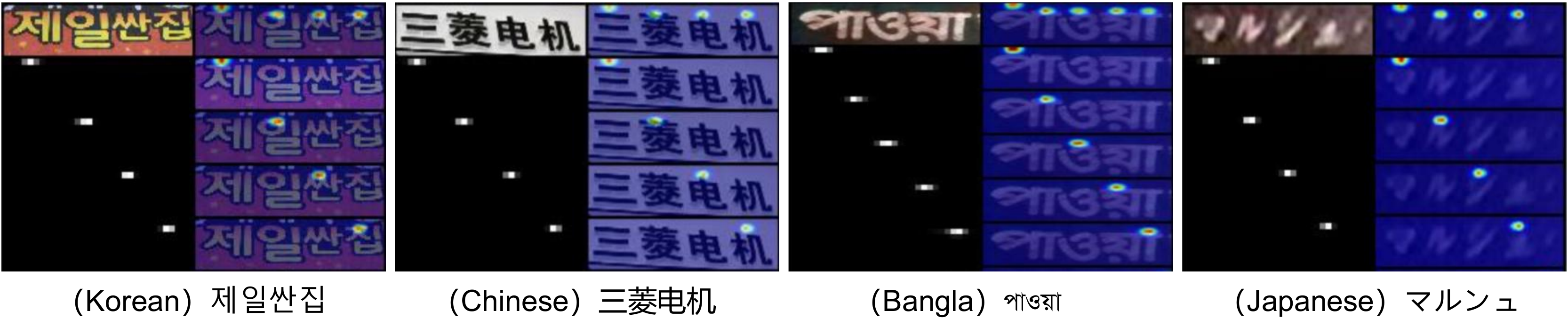} 
\caption{Step-by-step CDistNet visualization on MLT17 and MLT19. The heatmaps are visualized based on the position embedding after the third MDCDP.}
\label{vis_MLT_char}
\end{figure*}

The results are explained as follows. First, CDistNet and S-GTR both use the training data solely for semantic modeling. S-GTR can be viewed as an LM-style model that employs an additional textual reasoning module (here a graph neural network instead) for semantic reinforcement. On the contrary, CDistNet models the semantic information using the cross-attention-based query. Thus, the performance gains are mostly attributed to the more advanced feature modeling scheme, i.e., MDCDP. When comparing ABINet and CDistNet, ABINet applies the position embedding to visual and semantic in sequence. In the semantic space, it generates a roughly aligned character sequence at first then emphasizes the use of vocabulary for character-level error correction, where a strong LM constructed using Wikipedia data is leveraged for ensuring accuracy. We argue that in such a scheme visual and semantic clues are loosely related, which is prone to causing mistaken corrections in challenging scenarios. In Fig.\ref{vis_art} we list six recognition examples for further illustration. In five of them, the green marked characters are frequently used words, which guides ABINet in making incorrect recognition. On the other hand, by exploring recognition clues from the two domains and tying them together, CDistNet gets features better synergies these clues and achieves a more robust recognition. In Fig.\ref{vis_char}, we visualize the step-by-step attention heatmaps of three instances recognized by CDistNet. It is seen that features and characters are mostly well-aligned.

\begin{table*}{}
\setlength{\tabcolsep}{3pt}
\centering
\caption{Quantitative comparison of different methods on HA-IC13 (left) and CA-IC13 (right). H1 denotes HA-IC13 with stretching intensity 1. The others are defined similarly.}
\begin{tabular}{c|c|cccccc|cccccc}
\hline 
Method & Raw & HA1 & HA2 & HA3 & HA4 & HA5 & HA6 & CA1 & CA2 & CA3 & CA4 & CA5 & CA6\\
\hline
Transformer* \citep{sheng2019nrtr} & \textbf{97.2} & 96.3 & 95.5 & 92.4 & 86.5 & 79.4 & 72.5 & 95.7 & 94.4 & 85.9 & 75.9 & 65.9 & 58.6\\
RobustScanner* \citep{yue2020robustscanner}& 96.9 & 96.2 & 95.3 & 93.2 &  88.9 & 81.1 & 71.5 & 95.2 & 94.9 & 85.3 & 76.6 & 68.4 & 60.8 \\
S-GTR \citep{he2021S-GTR} & 93.8 & 91.4 & 91.2 & 86.1 & 79.7 & 68.5 & 58.8 & 91.4 & 89.0 & 81.4 & 70.7 & 61.0 & 53.3  \\
VisionLAN \citep{wang2021FTO} & 96.3 & 93.6 & 92.9 & 90.0 & 82.3 & 72.2 & 61.0 & 94.9 & 92.8 & 84.0 & 75.0 & 64.3 & 52.7  \\
ABINet \citep{ABInet21CVPR} & 97.0 & 95.9 & 95.2 & 92.0 & 85.8 & 73.8 & 65.0 & \textbf{96.6} &\textbf{95.9} & 87.9 & 76.3 & 65.5 & 54.5 \\
\hline
CDistNet w/o TPS & 97.1 & 96.4 & 96.1 & \textbf{94.5} & 89.9 & 83.0 &  76.1 & 96.3 & 94.9 & \textbf{88.6} & 78.3 & 70.1 &62.2 \\
CDistNet & 97.1 & \textbf{96.6} & \textbf{96.2} & 94.3 & \textbf{90.0} & \textbf{83.4} &  \textbf{77.7} & 96.3 & 95.6 & 88.5 & \textbf{79.6} & \textbf{70.4} &\textbf{63.1} \\
\hline
\end{tabular}
\label{table:HA_CA Ablation}
\end{table*}

\subsection{Experiments on MLT17 and MLT19}
We then experiment on MLT17 and MLT19 datasets for evaluating these methods on multilingual text recognition. We first encounter the problem that the language categories of the two datasets are not fully in accordance with our daily used ones. For example, they use Latin (including but more than English) rather than English. Therefore, it may not be easy to obtain readily available large-scale external LMs for each language. On the contrary, it also creates a venue to evaluate different methods in an LM-restricted manner. That is, the LM is merely constructed based on the training data of this language, which is relatively limited in quantity and thus the obtained LM is not powerful enough in general. We term this scenario as poor linguistic support in the paper.

Quantitative results on the two datasets are given in the right half of Table \ref{tab:art_uber_mlt}. The results are obtained by averaging the accuracy on seven (MLT17) or eight (MLT19) different languages. Similar to the above, CDistNet gains clear accuracy advantages over S-GTR and ABINet. It outperforms S-GTR by 3.37\% and ABINet by 3.29\% (corresponding to 5.29\% and 5.16\% relative improvements) on MLT17, and S-GTR by 3.53 and ABINet by 4.48\% (corresponding to 5.59\% and 7.2\% relative improvements) on MLT19. 

It is observed that ABINet does not show advantages to S-GTR. This can be explained from the LM usage perspective. Since we do not have a strong LM. Therefore ABINet, a more LM-dependent model, performs not as expected. While S-GTR can use a graph-based reinforcement to compensate for degradation caused by the weak LM somewhat, therefore eliminating their differences. To verify this, six illustrative examples are given in Fig.\ref{vis_MLT}. It is seen that only the Bangla example benefits from LM for ABINet. While S-GTR can give slightly better results directly (e.g., in Chinese). On the contrary, owing to rich associations in and between visual and semantic domains, feature-character alignments are still well established for CDistNet, as the examples shown in Fig.\ref{vis_MLT_char}. Therefore, it maintains the performance advantages and successfully recognizes all the instances. It verifies that CDistNet is less sensitive to the LM built on small datasets, again demonstrating the robustness of modeling multi-domain character distance.

\begin{figure*}[]
\centering
\includegraphics[width=1\textwidth]{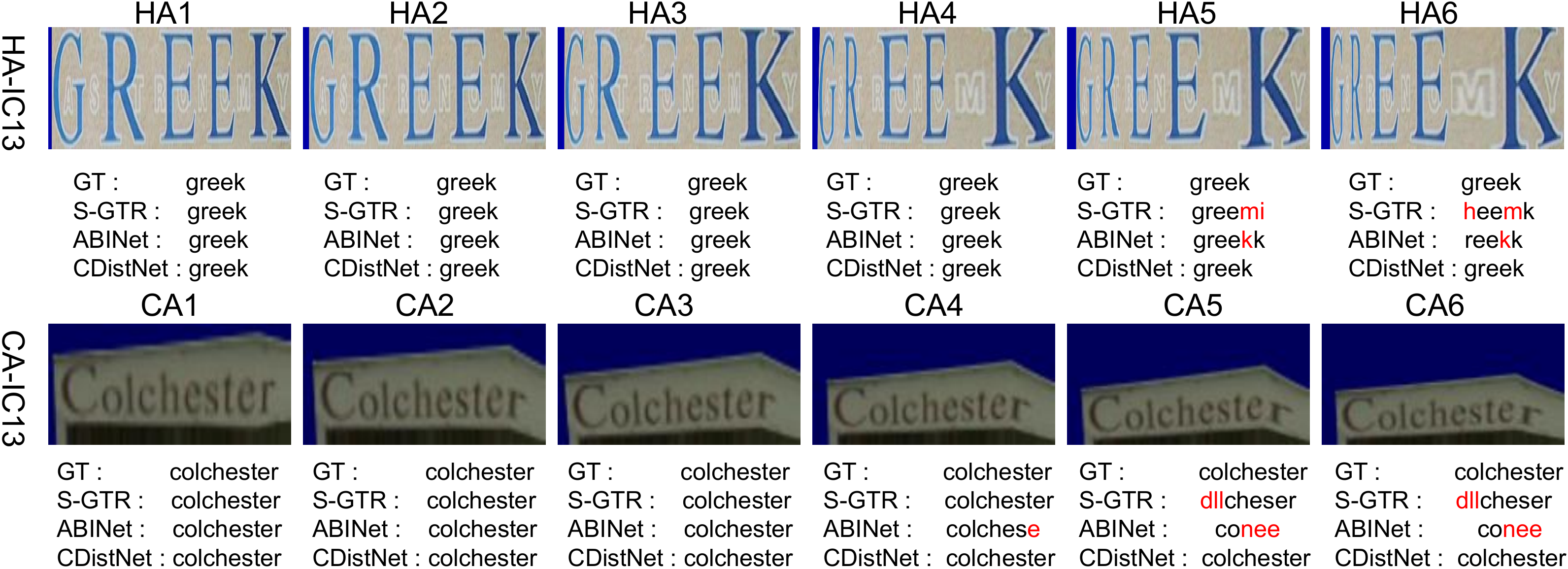} 
\caption{Recognition examples from HA-IC13 and CA-IC13 datasets. From left to right each example is rendered with more severe deformation and exhibits larger character distance variants. Incorrectly recognized characters are marked in red.}
\label{vis_haca}
\end{figure*}

\subsection{Experiments on HA-IC13 and CA-IC13}
Through the subjective assessment above, HA-IC13 and CA-IC13 form scene text recognition models evaluation testbeds with increasing difficulties. In addition, since the character shape is still largely maintained even though the text undergoes a large stretching intensity level, the two datasets are venues for assessing text recognizers with varying character layouts.

We validate CDistNet in both series of datasets and Tab.\ref{table:HA_CA Ablation} presents the results, where ABINet \citep{ABInet21CVPR}, VisionLAN \citep{wang2021FTO} and S-GTR \citep{he2021S-GTR} are three representative models in standard benchmarks. All methods recognize well in raw IC13 due to its simplicity. With the rise of deformation intensity, CDistNet shows its superiority. It goes down slower. The accuracy gaps gradually become larger with the intensity level increase, where margins ranging from 12.7\% to 18.9\%, and from 8.6\% to 10.4\% are respectively observed compared with the three methods in HA6 and CA6. The remarkable gap clearly verifies the great generalization ability of CDistNet especially in difficult scenarios. In Figure \ref{vis_haca}, we present two examples respectively on HA-IC13 and CA-IC13. As can be seen, the argumentation generates unequally spaced and severely bent text, corresponding to diverse character layouts. The recognition results of S-GTR, ABINet and CDistNet at different intensity levels are also given. All methods correctly recognize the two examples in low-intensity cases. With the recognition difficulties increasing, rarely-seen character layouts appear. S-GTR and ABINet begin to make mistakes. While CDistNet still robustly recognizes them. It shows better robustness to rare character layouts, as it well perceives and integrates multi-domain character distances.

We also observe that the sequential (autoregressive) decoding methods behave differently from the parallel decoding ones from this experiment. VisionLAN, ABINet and S-GTR, which employ the parallel decoding scheme that decodes all characters in a single forward pass for speed acceleration, experience even sharp accuracy decreases compared to the three sequential decoding methods, i.e., NRTR*, RobustScanner* and CDistNet that decodes the characters one-by-one. For example, CDistNet is better than VisionLan, ABINet and S-GTR by 16.7\%, 12.7\% and 18.9
\% in HA6, 10.4\%, 8.6\% and 9.8\% in CA6. NRTR* and RobustScanner* also have large margins compared to them. The significant accuracy gaps can be understood as different models make different tradeoffs between accuracy and speed. The parallel decoding scheme hardly constructs a decoding context as fine as the sequential decoding scheme. Their fast speed is established at the cost of sacrificing the recognition quality in difficult text. Note that we are the first to observe the ineffectiveness of popular parallel decoding schemes in handling such difficult text. Tackling this drawback should be an interesting topic worthy of further study.

\section{Conclusion}
Targeting to improve the accuracy of scene text recognition especially in difficult scenarios, we have presented the MDCDP module which utilizes the position feature to \emph{query} both visual and semantic features within the Transformer-based encoder-decoder framework. It learns a comprehensive and robust feature representation that delineates character distance in both visual and semantic domains, and more importantly, ties the two domains together. Accordingly, CDistNet has been developed for robust scene text recognition. We have conducted extensive experiments to verify its effectiveness. While ablation studies demonstrate the effect of each proposed component. The comparison results successfully validate our proposal. It reports state-of-the-art accuracy when compared with existing methods on the six standard benchmarks. Moreover, on larger and more challenging real-world datasets and augmented datasets presenting different recognition challenges, it also shows clear advantages compared to methods that perform quite similarly in the standard benchmarks. In addition, the visualization experiments verify that superior attention localization and reasonable semantic utilization are reached in visual and semantic domains, respectively. 

It is observed that the accuracy on the six standard benchmarks is near saturation, which might be insufficient to well distinguish the methods performed top-tier. Future work should be more emphasize larger and more challenging datasets, not only ArT, Uber, MLT17 and MLT19, but also the created HA-IC13 and CA-IC13. We hope these datasets can accelerate future research in designing robust scene text recognition models. Meanwhile, accuracy and inference speed are always two key metrics for scene text recognition. A recent study reported effective yet efficient solutions by leveraging ViT-like architecture~\citep{Du2022SVTR}. We plan to incorporate it to speed up the recognition. In addition, we are also interested in developing more robust models for text spotting \citep{fang2022abinet++} and page-level text recognition \citep{peng2022pagenet}.

\begin{acknowledgements}
This work was supported by the National Natural Science Foundation of China under Grants 62172103 and 62102384.
\end{acknowledgements}

\bibliographystyle{spbasic}      
\bibliography{egbib}   

\end{document}